\documentclass{article}

\usepackage{PRIMEarxiv}

\usepackage[utf8]{inputenc} 
\usepackage[T1]{fontenc}    
\usepackage{hyperref}       
\usepackage{url}            
\usepackage{booktabs}       
\usepackage{longtable,array} 
\usepackage{caption}
\usepackage{amsfonts}       
\usepackage{nicefrac}       
\usepackage{microtype}      
\usepackage{lipsum}
\usepackage{fancyhdr}       
\usepackage{graphicx}       
\usepackage{cleveref}
\usepackage{listings}
\usepackage{bera}
\usepackage{listings}
\usepackage{xcolor}
\graphicspath{{media/}}     

\colorlet{punct}{red!60!black}
\definecolor{background}{HTML}{EEEEEE}
\definecolor{delim}{RGB}{20,105,176}
\colorlet{numb}{magenta!60!black}

\lstdefinelanguage{json}{
    basicstyle=\normalfont\ttfamily,
    numbers=left,
    numberstyle=\scriptsize,
    stepnumber=1,
    numbersep=8pt,
    showstringspaces=false,
    breaklines=true,
    frame=lines,
    backgroundcolor=\color{background},
    literate=
     *{0}{{{\color{numb}0}}}{1}
      {1}{{{\color{numb}1}}}{1}
      {2}{{{\color{numb}2}}}{1}
      {3}{{{\color{numb}3}}}{1}
      {4}{{{\color{numb}4}}}{1}
      {5}{{{\color{numb}5}}}{1}
      {6}{{{\color{numb}6}}}{1}
      {7}{{{\color{numb}7}}}{1}
      {8}{{{\color{numb}8}}}{1}
      {9}{{{\color{numb}9}}}{1}
      {:}{{{\color{punct}{:}}}}{1}
      {,}{{{\color{punct}{,}}}}{1}
      {\{}{{{\color{delim}{\{}}}}{1}
      {\}}{{{\color{delim}{\}}}}}{1}
      {[}{{{\color{delim}{[}}}}{1}
      {]}{{{\color{delim}{]}}}}{1},
}
\usepackage[title]{appendix}

\newcommand{\benchmark}{EU AI Act Q\&A Benchmark}
\newcommand{\challenge}{2026 EU AI Act Q\&A Challenge}

\title{Are LLM Assistants Regulatory Ready?\\Answers from regenold's 2026 EU AI Act Q\&A Challenge
}

\author{
  Marco Virgolin \\
  Unlayer AI / regenold GmbH \\
  Amsterdam, the Netherlands \\
  \texttt{marco@\{unlayer.ai/regenold.com\}} \\
    \And
  Thomas Haszprunar \\
  regenold GmbH \\
  Badenweiler, Germany \\
  \texttt{thomas.haszprunar@regenold.com} \\
   \And
  Maximilian Regenold \\
  regenold GmbH \\
  Badenweiler, Germany \\
  \texttt{max@regenold.com} \\
}

\begin{document}
\maketitle

\begin{abstract}
Large language model (LLM) assistants are increasingly proposed for regulatory affairs, yet their reliability in regulatory tasks remains unclear.
We introduce the \benchmark{}, an expert-curated benchmark of 110 open-ended questions about the EU AI Act (\emph{Regulation (EU) 2024/1689}).
Each question comes with an exemplary answer, question-specific correctness criteria, and expected references.
Evaluation measures answer correctness, reference quality, conciseness, tone, and latency, with a multi-turn ``hard mode'' testing responses to simulated user pushback.
We also present the results of the \challenge{}, based on the benchmark, where three off-the-shelf baselines and five purpose-built assistants (from four commercial developers and one academic entrant) were tested.
Only two of the five submitted assistants consistently outperformed the baselines; one achieved high scores across all evaluated metrics.
These results show that specialist development alone does not guarantee better performance and highlight the critical importance of LLM assistant evaluation in regulatory affairs.
The benchmark code is openly available at \url{https://github.com/regenold-GmbH/eu-ai-act-benchmark-2026/}.
\end{abstract}


\section{Introduction}

Large language models (LLMs) are becoming increasingly capable in many areas, including tasks that pertain to regulatory processes, such as document review, summarization and drafting, regulatory research, and question answering.
However, developing reliable LLM assistants for regulatory affairs remains particularly challenging because they must navigate complex, interconnected requirements across a large and evolving body of regulations and guidance.
Addressing a single regulatory question may require cross-referencing international standards, Union law, and national guidance, all of which must be kept up to date.  
Furthermore, in sectors such as medicinal product or medical device development, regulatory mistakes carry costs that range from heavy to unacceptable, including delayed access to treatment and harm to patients.

In recent years, many benchmarks have been proposed to evaluate the performance of LLMs---intended as text comprehension and generation models---and LLM assistants---intended as \emph{agents}, where software surrounding the model enables iterative observation, reasoning, and action---across a variety of disciplines.
Regarding legal and regulatory affairs, modern benchmarks cover reasoning, retrieval, compliance assessment, and question answering (see \Cref{sec:related}).
For organizations considering to adopt regulatory LLM assistants, however, a practical question often remains: do purpose-built  systems produce better answers than off-the-shelf ones on realistic tasks?

To address this question, we introduce a benchmark and present the results of a challenge where commercial and academic contestants submitted their purpose-built assistants, focused on the EU Artificial Intelligence Act (Regulation (EU) 2024/1689).
The AI Act was chosen due to its broad scope and cross-sector relevance: it applies across industry and societal sectors, and affects system providers, organizations deploying these systems, and individuals exposed to them.
Our contributions are as follows.
First, we introduce our expert-curated \benchmark{}, with question-level correctness criteria, exemplary answers, expected references, and other meta-data among which estimated difficulty.
Second, we present the respective evaluation pipeline, that includes deterministic checks and LLM-as-judge scoring, for Q\&A with and without a simulated multi-turn setup that includes an adversarial question.
Third, we report the results of the \challenge{}, in which five purpose-built LLM assistants and three off-the-shelf baselines were evaluated on the benchmark without prior sight of the questions, to assess their readiness for regulatory question answering.

Now that the \challenge{} concluded, we release the benchmark and evaluation pipeline code so that others can evaluate their own systems: \url{https://github.com/regenold-GmbH/eu-ai-act-benchmark-2026/}.

\section{Related Work}
\label{sec:related}
\paragraph{Legal and regulatory benchmarks.}
The class of benchmarks that is most related to the one proposed here is legal and regulatory benchmarks.
Legal-domain natural language processing (NLP) benchmarks have moved from general understanding tasks toward reasoning, retrieval, and practical work-product evaluation.
LexGLUE collects English legal language-understanding datasets in a standardized benchmark
\cite{chalkidis2022lexglue}. 
LegalBench is a collaboratively built benchmark
of 162 legal reasoning tasks authored with legal experts \cite{guha2023legalbench}. 
LegalBench-RAG extends this direction toward legal-retrieval by retracing LegalBench contexts to source locations and emphasizing minimal relevant snippets for retrieval-augmented generation
\cite{pipitone2024legalbenchrag}. 
More recent legal-retrieval work focuses on reasoning over legal authorities rather than only lexical matching \cite{zheng2025legalretrieval}.

Each question in our benchmark has a checklist of question-specific correctness criteria.
Legal benchmarks and leaderboards have also begun to evaluate AI systems with task-specific rubrics.
Harvey's Legal Agent Benchmark, for example, uses matter-centric tasks and expert rubrics for long-horizon legal
agent work~\cite{harvey2026lab};
Legal AI Benchmarking reports substantive task success and output form separately for tasks such as extraction and contract drafting
\cite{legalbenchmarks2026leaderboard}.
These examples motivate rubric-based evaluation of complex workflows.

Examples of benchmarks focusing on regulatory question answering include ObliQA and ObliQA-MP, which evaluate retrieval and answer generation over financial regulations~\cite{gokhan2024rirag,gokhan2025obliqamp}, and FDARxBench, which evaluates document-grounded questions motivated by FDA generic-drug assessment, using expert-guided, LLM-assisted question generation and passage-level provenance~\cite{xiong2026fdarxbench}.

A few benchmarks focus specifically on the EU AI Act.
AIReg-Bench evaluates how closely LLM assessments of AI Act compliance match those of legal experts, using synthetic technical-documentation excerpts~\cite{marino2025aireg}.
Tsalera et al.\ evaluate AI Act comprehension using 100 multiple-choice questions~\cite{tsalera2026aiact}.
Davvetas et al.\ provide tasks for risk classification, article retrieval, obligation generation, and question answering~\cite{davvetas2026aiactbenchmark}.
Our work differs from the above in that it proposes expert-curated, open-ended question answering that is measured comprehensively, i.e., across answer correctness, reference quality, conciseness, tone, and latency, under single-turn and multi-turn settings; further, we compare five independently submitted purpose-built assistants with three off-the-shelf baselines, via a dedicated challenge.

\paragraph{Citation quality.}
In legal or regulatory affairs, an answer is only of use if the user can trace it back to the law or regulation it rests on: a claim that cannot be verified against the source is not reliable.
Hallucination is therefore a central risk for LLM use in law. 
Dahl et al.\ document that models can produce false legal claims and fail to correct false premises, warning against unsupervised deployment in legal settings \cite{dahl2024legalfictions}.
Citation-oriented benchmarks such as ALCE evaluate whether generated answers are supported by citations~\cite{gao2023alce} and legal-retrieval benchmarks argue that precise source retrieval is crucial for verifiable downstream answers \cite{pipitone2024legalbenchrag}. 
Our benchmark follows this motivation but uses the form of the answer itself as the citation target: systems must return normalized Article and Annex references, which are scored separately from textual answer correctness.

\paragraph{LLM-as-judge and multi-metric evaluation.}
LLM-as-judge (here also referred to as ``LLM judge'') methods are widely used for open-ended generation tasks because string overlap metrics are poorly suited to answers where many wordings can be correct. 
G-Eval showed that \texttt{gpt-4}-style rubric evaluation can better align with human assessments for generation tasks than traditional metrics \cite{liu2023geval}.
Zheng et al. studied LLM judges using MT-Bench and Chatbot Arena for multi-turn chat evaluation and documented both practical usefulness and biases such as position, verbosity, and self-enhancement effects \cite{zheng2023mtbench}.
Prometheus further showed that models can be trained for fine-grained rubric evaluation with custom criteria \cite{kim2024prometheus}. 
Our evaluation uses this rubric-driven paradigm, but constrains the LLM judge to binary pass/fail decisions over question-specific correctness criteria and repeats judge calls to expose variability.

Our benchmark also follows the multi-metric philosophy of HELM, which argues that language-model evaluation should report more than accuracy and should surface trade-offs such as robustness and efficiency \cite{liang2023helm}.
Our metrics separate substantive correctness, reference quality, verbosity, tone, and speed so that a system cannot hide a serious weakness behind a single aggregate score.
The hard modality is also related to recent multi-turn evaluation work such as MultiChallenge, which tests whether frontier models retain instructions and reason reliably across realistic conversation histories
\cite{sirdeshmukh2025multichallenge}.

\section{The Benchmark}

\subsection{Scope}
The benchmark covers Regulation (EU) 2024/1689 of 13 June 2024, interpreted against the state of affairs as of 1 May 2026. Subsequent amendments, corrigenda, delegated acts, implementing acts, guidance, or interpretations are outside the stated scope of the current evaluation.
For example, the \emph{AI Omnibus}~\cite{eu2026aiomnibus}, which entered into force on July 27, 2026 and includes new deadlines as well as prohibitions (e.g., nudifier systems), is out of scope.

\subsection{Dataset Construction}
\label{sec:dataset-construction}
The dataset was created primarily through expert authoring and review. 
Limited automation was used for question discovery. 
Namely, \texttt{gpt-5.4-mini} was used to create questions out of random chunks of the legislation, and these questions were manually adjusted or discarded. 
The dataset contains 110 \emph{items}, each composed of the following fields:
\begin{itemize}
    \item Question: the question.
    \item Answer: an exemplary answer to the question.
    \item Correctness criteria: salient aspects that should be present and correct in the candidate answer.
    \item References: a comma-separated list of Articles or Annexes that should be cited as sources to answer the question.
    \item Difficulty: the annotator's estimated difficulty of the question (easy, medium, hard).
    \item Text excerpt (optional): a text chunk from the act concerning the question.
    \item Author: the person who authored the item.
    \item Notes: any custom notes.
\end{itemize}

Example items are provided in \Cref{tab:example-benchmark-items}.

\begin{table*}[t]
\caption{Question, correctness criteria, and expected references for three exemplary benchmark items,
illustrating increasing complexity, from a single-definition question to a multi-actor application scenario.}
\label{tab:example-benchmark-items}
\centering
\small
\setlength{\tabcolsep}{4pt}
\begin{tabular}{@{}>{\raggedright\arraybackslash}p{0.28\linewidth}>{\raggedright\arraybackslash}p{0.39\linewidth}>{\raggedright\arraybackslash}p{0.18\linewidth}>{\raggedright\arraybackslash}p{0.08\linewidth}@{}}
\toprule
Question & Correctness criteria & Expected references & Difficulty \\
\midrule
Under Regulation (EU) 2024/1689 (EU AI Act), how is ``risk'' defined?
& \textbf{(1)} Combination of probability and severity of harm
& Article 3.2
& Easy \\
\addlinespace[5pt]
What are the conditions to classify a general-purpose AI (GPAI) model as a ``GPAI model with systemic risk''? Do all need to be met at the same time?
& \textbf{(1)} Condition: High-impact capabilities\par
  \textbf{(2)} Presumption: high-impact if training compute $>10^{25}$ FLOPs\par
  \textbf{(3)} Condition: Commission can designate ex officio (e.g. after scientific-panel alert)\par
  \textbf{(4)} Either confidition suffice.
& Article 51
& Medium \\
\addlinespace[5pt]
A private contractor runs a region's gas supply service on behalf of the local government. It licenses an AI system from a vendor, fine-tunes it on local data, and then puts it into service under its own brand. To learn about the correct classifications of the system and contractor, what material of the AI Act should be read? Does a fundamental rights impact assessment (FRIA) apply?
& \textbf{(1)} Article 6 and Annex III p.2 to understand if AI system is high-risk\par
  \textbf{(3)} Article 25.1 to understand if contractor is provider\par
  \textbf{(4)} Articles 16 and 26: obligations of providers and deployers\par
  \textbf{(5)} FRIA does not apply to Annex III p.2 uses
& Article 6.2; Annex III.2; Article 25.1; Article 16; Article 26; Article 27.1
& Hard \\
\bottomrule
\end{tabular}
\end{table*}

\begin{figure}
    \centering
    \includegraphics[width=1.0\linewidth]{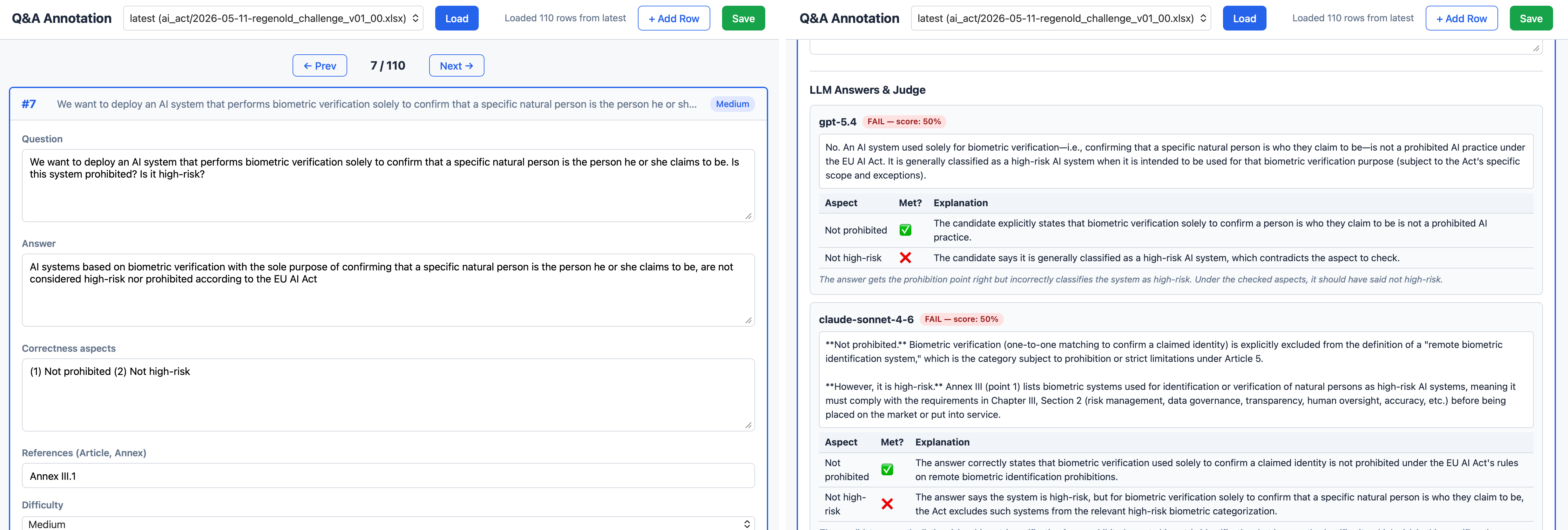}
    \caption{Screenshots of the UI used to annotate the dataset. The left part shows the top of the UI, where the annotator would enter several key fields. The right part shows a later section of the UI, where the annotator could use off-the-shelf models to assess how the question is answered and scored by the LLM-as-a-judge with respect to the correctness criteria.}
    \label{fig:annotation-UI}
\end{figure}

Item authoring was supported by a custom-built user interface (UI), see~\Cref{fig:annotation-UI}.
Annotators could add, remove, and edit questions and related fields, but also generate trial answers and respective judgments from off-the-shelf LLMs; namely, the annotators could query OpenAI's \texttt{gpt-5.4}, Anthropic's \texttt{claude sonnet 4.6}, and \texttt{search-integrated LLM} to propose answers, while \texttt{gpt-5.4} was used to judge the criteria (all LLM calls using default parameters).
The central design choice was to author explicit correctness criteria rather than score by direct similarity to the reference answer.
Preliminary testing indicated that leveraging criteria produced better calibration between the LLM-as-a-judge and author than asking the judge to compare candidate answers directly to a prose reference answer.
Furthermore, iterative refinement informed by observed LLM responses helped ensure that questions and their assessment criteria accommodated differences in phrasing while consistently evaluating substantive content.
Up to five correctness criteria were authored per answer, to keep judging easy.

\begin{figure}[t]
    \centering
    \begin{tabular}{@{}ccc@{}}
        \includegraphics[width=0.32\linewidth]{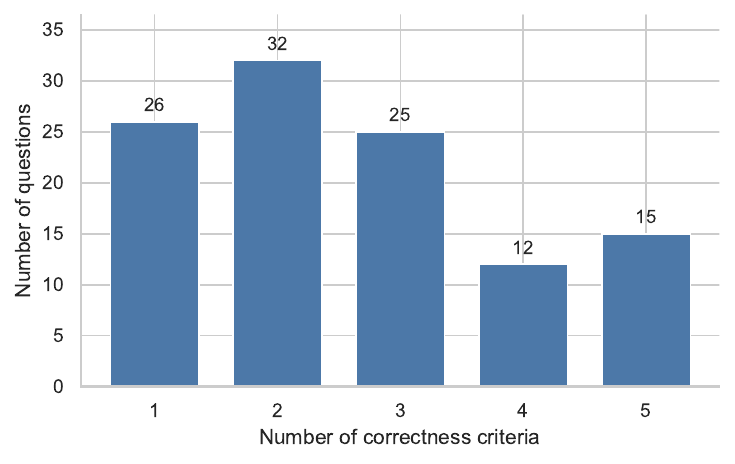} &
        \includegraphics[width=0.32\linewidth]{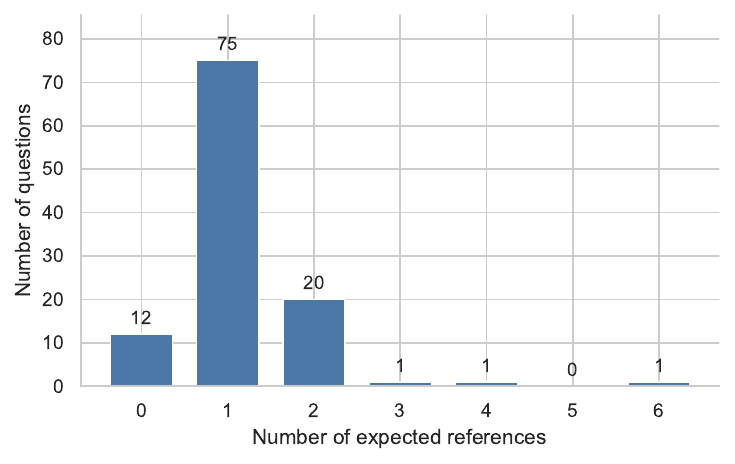} &
        \includegraphics[width=0.32\linewidth]{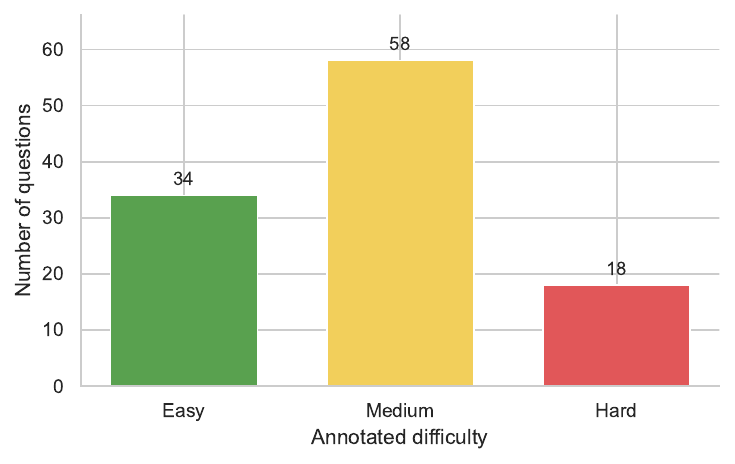} \\
    \end{tabular}
    \caption{Dataset distributions: number of correctness criteria per item, expected references per item, and annotator-perceived question difficulty.}
    \label{fig:dataset-distributions}
\end{figure}

Distributions of correctness criteria, expected references, and question difficulty, are available in \Cref{fig:dataset-distributions}.

\subsection{Evaluation Metrics}
\label{sec:evaluation-metrics}

The benchmark reports eight metrics and an overall score, summarized in \Cref{tab:metrics}. 

\begin{table}[t]
\centering
\caption{Benchmark metrics and their definitions.}
\label{tab:metrics}
\begin{tabular}{p{0.28\linewidth}p{0.62\linewidth}}
\toprule
\textbf{Metric} & \textbf{Definition} \\
\midrule
Answer correctness, loose & Percentage of individual correctness criteria satisfied by the candidate answer. \\
Answer correctness, strict & Percentage of questions for which all required correctness criteria are satisfied. \\
Answer conciseness & Inverted measure of answer verbosity relative to the reference answer. \\
Reference correctness, loose & Percentage of expected top-level Articles or Annexes found. \\
Reference correctness, strict & Percentage of expected references found when keeping the first Article/Annex sublevel. \\
Reference conciseness & Inverted measure of excess references, relative to the expected references. \\
Regulatory tone & Fraction of responses judged both professionally appropriate and clear for a regulatory context. \\
Response speed & Inverted measure of latency. \\
\midrule
Overall score & Geometric mean of the eight above. \\
\bottomrule
\end{tabular}
\end{table}

Formally, let item $i \in 1, \dots N$ have $m_i$ correctness criteria and binary outcomes $c_{ij}\in\{0,1\}$ (1=pass, 0=fail). 
Then the scores are defined as follows.

\emph{Loose answer correctness} is:
\[
100 \cdot
\frac{\sum_i \sum_{j=1}^{m_i} c_{ij}}{\sum_i m_i},
\]
and \emph{strict answer correctness} is:
\[
100 \cdot \frac{1}{N}\sum_i \mathbf{1}\left[\prod_{j=1}^{m_i} c_{ij}=1\right],
\]
where $c_{ij}$ are produced by an LLM-as-a-judge (details below).

\emph{Answer conciseness} is computed as 
\[
100 \cdot \frac{1}{N}\sum_i \frac{1}{\max\left(w_i/w_i^\star,1\right)},
\]
where $w_i$ and $w^\star_i$ are the number of words in the candidate and reference answer, respectively, and the max-with-one operation ensures that only excessive length is penalized.

For \emph{reference correctness}, let $E_i$ be the expected reference set and $F_i$ the candidate reference set. 
\emph{Loose} matching normalizes references to top-level Articles or Annexes (e.g., $Article 1$); \emph{strict} matching further considers the first sublevel (e.g., $Article 1.1$). If an expected reference has only a top-level and the expected reference matches it, but includes a sublevel, then strict matching succeeds (e.g., candidate is $Article 1.1$ and expected is $Article 1$).
Reference correctness is averaged over items as
\[
100 \cdot \frac{1}{N}\sum_i \frac{|E_i \cap F_i|}{|E_i|}.
\]
\emph{Reference conciseness} is computed similarly to answer conciseness, to be inversely proportional to the number of extra references:
\[
100 \cdot \frac{1}{N}\sum_i \frac{|E_i|}{|E_i|+|F_i\setminus E_i|}.
\]
A less-specific citation such as \texttt{Annex IV} is not counted as an unrelated extra when \texttt{Annex IV.1} was expected.
Some benchmark items expect no specific references to be cited: those are excluded from reference-correctness and conciseness averages.

\emph{Regulatory tone} is judged via LLM with a pass or fail, and averaged across number of items $N$.

Lastly, \emph{response speed} is calculated as a function of latency $\ell_i$ (in seconds), simply as:
\[
\frac{1}{N}\sum_i \max\left(100-\ell_i,0\right),
\]
meaning that every latency second reduces the score by one percentage point, and answers taking more than $100$ seconds score $0$.

An \emph{overall score} from 0 to 100\% is also computed, as the geometric mean of the eight metrics.
The geometric mean was chosen over e.g., the arithmetic mean, because even a single point of failure can jeopardize usability and trustworthiness in concrete regulatory workflows. 
For example, a system that answers always correctly (100\% correctness) but takes more than 100 seconds to respond (0\% speed) would obtain a score of 0, reflecting the fact that it is unusable. 

\paragraph{Scoring implementation} Pass/fail judgements for answer correctness and regulatory tone are evaluated by an LLM.
For answer correctness, the LLM receives the question, the candidate answer and the correctness criteria for that item; and is tasked to firstly emit a short consideration, and then return a pass/fail decision, for each criterion. 
For tone, the LLM receives the candidate answer and evaluates only whether the language is professional, formal, clear, unambiguous, and suitable for a regulatory context.
The prompts used to evaluate correctness and regulatory tone are provided in Appendix~\ref{apdx:sec:prompts}.
All other metrics were computed deterministically.

\subsection{Evaluation Protocol}

For each dataset item, the competing system receives a conversation history containing the question in exam (more specifics below), and must return a structured response with three fields:
\begin{itemize}
  \item an optional \texttt{reasoning} field, not used for scoring;
  \item an \texttt{answer} field containing the final answer text;
  \item a \texttt{references} list containing legal references.
\end{itemize}

References must follow a normalized legal format:
\texttt{Article NUMBER}, \texttt{Article NUMBER.SUBNUMBER},
\texttt{Annex ROMAN-NUMBER}, or \texttt{Annex ROMAN-NUMBER.SUBNUMBER}.
For example, \texttt{Article 3}, \texttt{Article 50.4}, and \texttt{Annex IV.2} are valid.

The benchmark has two modalities.
In \emph{easy mode}, the conversation history contains a single turn, with the \texttt{user} posing the question.
In \emph{hard mode}, the history begins with a fixed synthetic nine-turn conversation between the \texttt{user} and the   \texttt{assistant}, followed by the actual benchmark question as the final \texttt{user} message.
After the system responds, a synthetic user challenge is appended, stating that the answer may contain hallucinations and
inviting the system to reason before answering again.
The final answer after this adversarial pushback is the one that is scored.
This setup tests whether systems remain stable both with longer conversations---which are generally associated with worse performance~\cite{bai2024longbench,du2025context,liu2024lost}---and when a user challenges an answer that may already be   correct or partially correct---a situation in which LLMs may change their answers to agree with the user's expressed belief,   even when doing so produces an incorrect answer~\cite{ibrahim2026training,hong2025measuring,sharma2024towards}.
The synthetic history and the adversarial pushback are reported in Appendix~\ref{apdx:sec:hard-mode}.

\section{Experimental Setup: regenold’s 2026 EU AI Act Benchmark Challenge}

\paragraph{The challenge}
The experimental setup was realized in the form of a challenge, namely regenold’s \challenge: \url{https://regenold.com/landing/ai/eu-ai-act-competition}.

Contestants were given the general background of the challenge, the scope and evaluation metrics in general terms (e.g., ``answer correctness'', ``reference correctness'', ``latency'', etc.), the expected API format, and three example questions (listed in Appendix~\Cref{apdx:sec:example-questions}).
No benchmark items were provided and contestants were not informed of one another.

Approximately 50 entities, including companies, university labs, and individuals, were contacted, and five agreed to participate.
The five contestants enrolled between May and July 2026, with benchmark evaluations conducted on a first-come, first-served basis until approximately mid-August 2026.

Once all contestants were evaluated, an individual report was produced and shared with each separately, presenting the contestant's result against the two web search-based baselines (described below).
Based on the outcome in the individual report, contestants could contact us to have their entry anonymized. 
The contestants were:
\begin{itemize}
    \item Pharos Labs: Company, AI for healthcare (\url{https://www.pharos-labs.com});
    \item Ansvar AI: Company, AI tools for law and security (\url{https://www.ansvar.eu})
    \item Antifragile AI: Company, AI for compliance (\url{https://antifragile-ai.net}); 
    \item Anonymous 1 --- Company, opted for anonymization;
    \item Anonymous 2 --- University Lab, opted for anonymization.
\end{itemize}

\paragraph{Baselines}
Three systems served as baselines, using ``off-the-shelf'' APIs and libraries.
The first two relied on web search:
OpenAI's \texttt{gpt-5.5} (released in Q2 2026), equipped with the web search tool available through the OpenAI API (``\texttt{gpt-5.5} + web search''), and an undisclosed system\footnote{Major US company not providing explicit permission of publication of derived results in their terms.} which natively integrates web search (``\texttt{search-integrated LLM}'' released in Q1 2025).
A third baseline was implemented using basic retrieval augmented generation.
Namely, \texttt{gpt-5.5} was equipped with LlamaIndex to retrieve five excerpts and answer accordingly (``\texttt{gpt-5.5} + rag''); more details on this are in Appendix~\ref{apdx:sec:rag-baseline}.

\paragraph{LLM-as-a-judge}
The LLM-as-a-judge was implemented using \texttt{gpt-5.4}.
Judge calls were repeated three times at a temperature of 0.1 to assess variation in scoring.
Post competition, a verification of the calibration level of the LLM-as-a-judge was conducted:
Fifty answer judgments were manually reviewed, sampling uniformly at random between \emph{informative} cases---i.e., cases where the LLM-as-a-judge consistently scored a criterion as failed \emph{or} scored differently across the three repetitions---and \emph{ordinary} cases---uniform random case sampling.

\section{Results}
This section reports the \challenge{} results for all contestants and the three baselines, first summarizing overall performance and then breaking down results by evaluation metric and difficulty setting. 
Common error patterns and systematic failure modes that emerged across systems are also presented.

\begin{figure*}[t]
\centering
\includegraphics[width=0.92\textwidth]{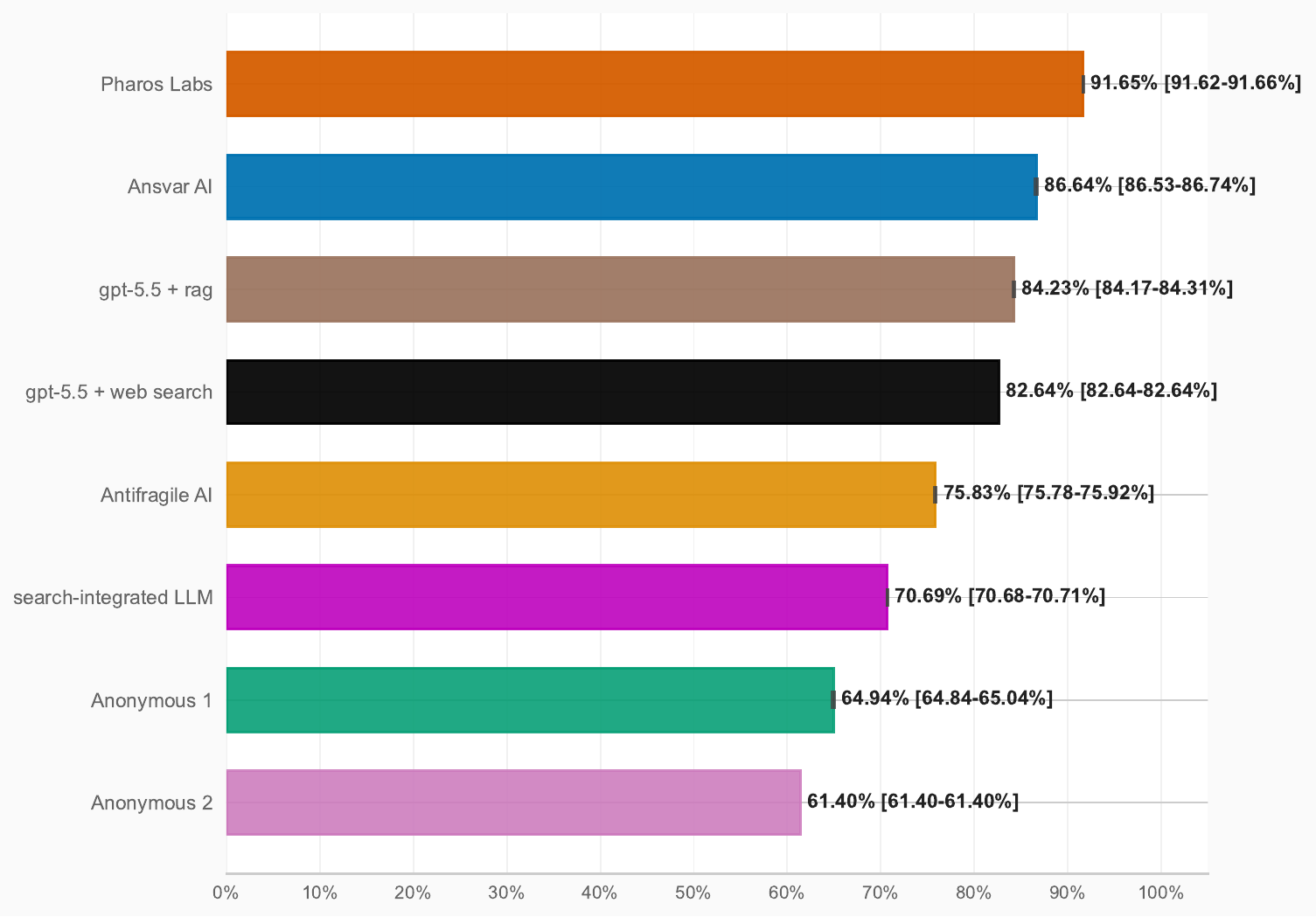}
\caption{Easy-mode overall scores. Horizontal whiskers and bracketed labels show the observed minimum--maximum range across the three judge runs.}
\label{fig:overall-score}
\end{figure*}

\begin{figure*}[t]
\centering
\includegraphics[width=0.92\textwidth]{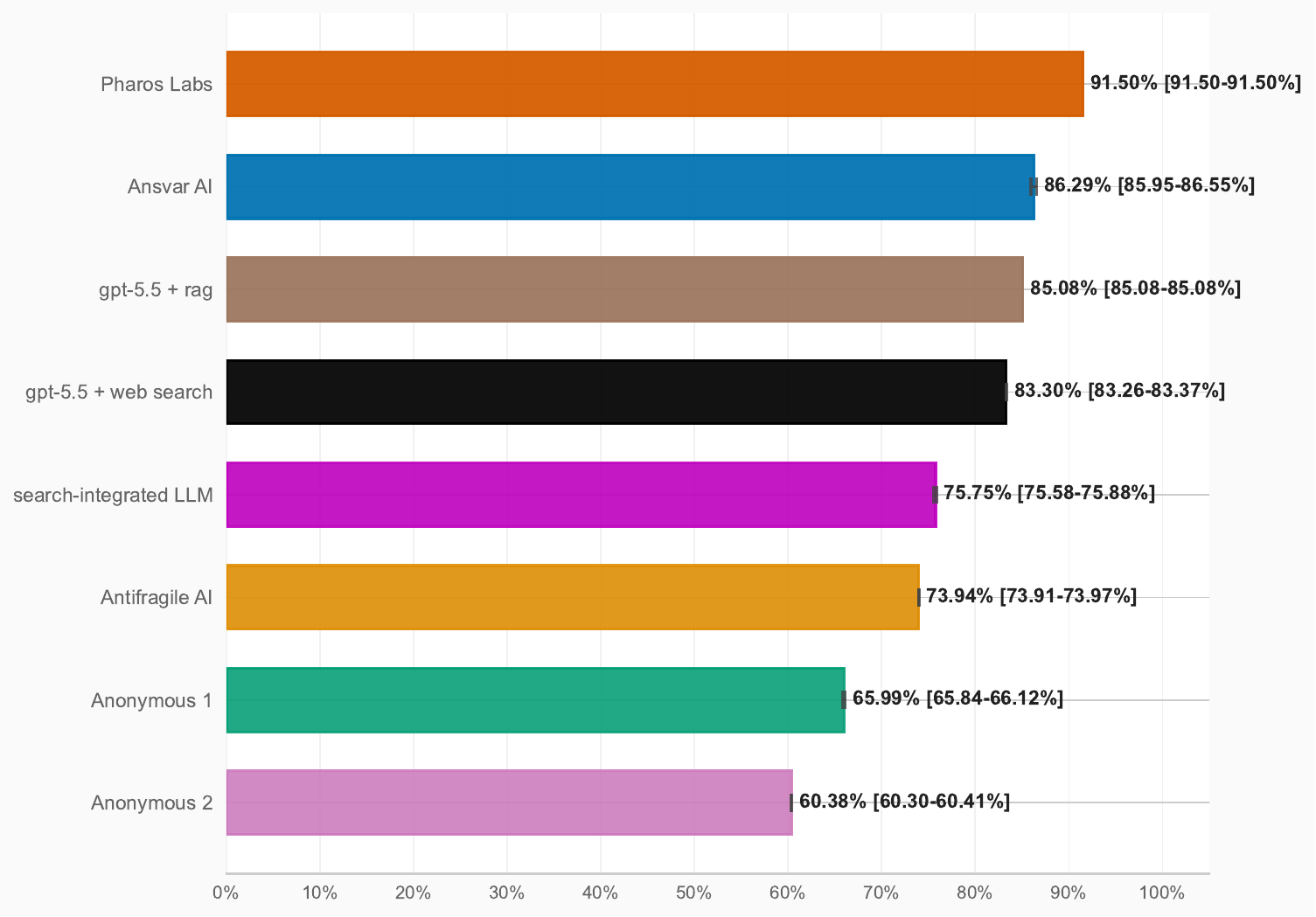}
\caption{Hard-mode overall scores. Horizontal whiskers and bracketed labels show the observed minimum--maximum range across the three judge runs.}
\label{fig:overall-score-hard}
\end{figure*}

\paragraph{Overall scores}
\Cref{fig:overall-score,fig:overall-score-hard} visualize the overall scores of the contestants and baselines.
Reported numbers are mean scores where the LLM-as-a-judge scoring was involved, with minimum--maximum scores shown in brackets.
Pharos Labs achieved the highest overall score in both easy and hard mode, followed by Ansvar AI.
These were the only two contestants to outperform all baselines in both modes.
Next, \texttt{gpt-5.5} + rag and web search ranked as third and fourth, respectively, in both modes.
Antifragile AI ranked fourth in easy mode, ahead of the \texttt{search-integrated LLM} baseline, but fell just below that baseline in hard mode.
The anonymous contestants followed.

\begin{figure*}[t]
\centering
\includegraphics[width=\textwidth]{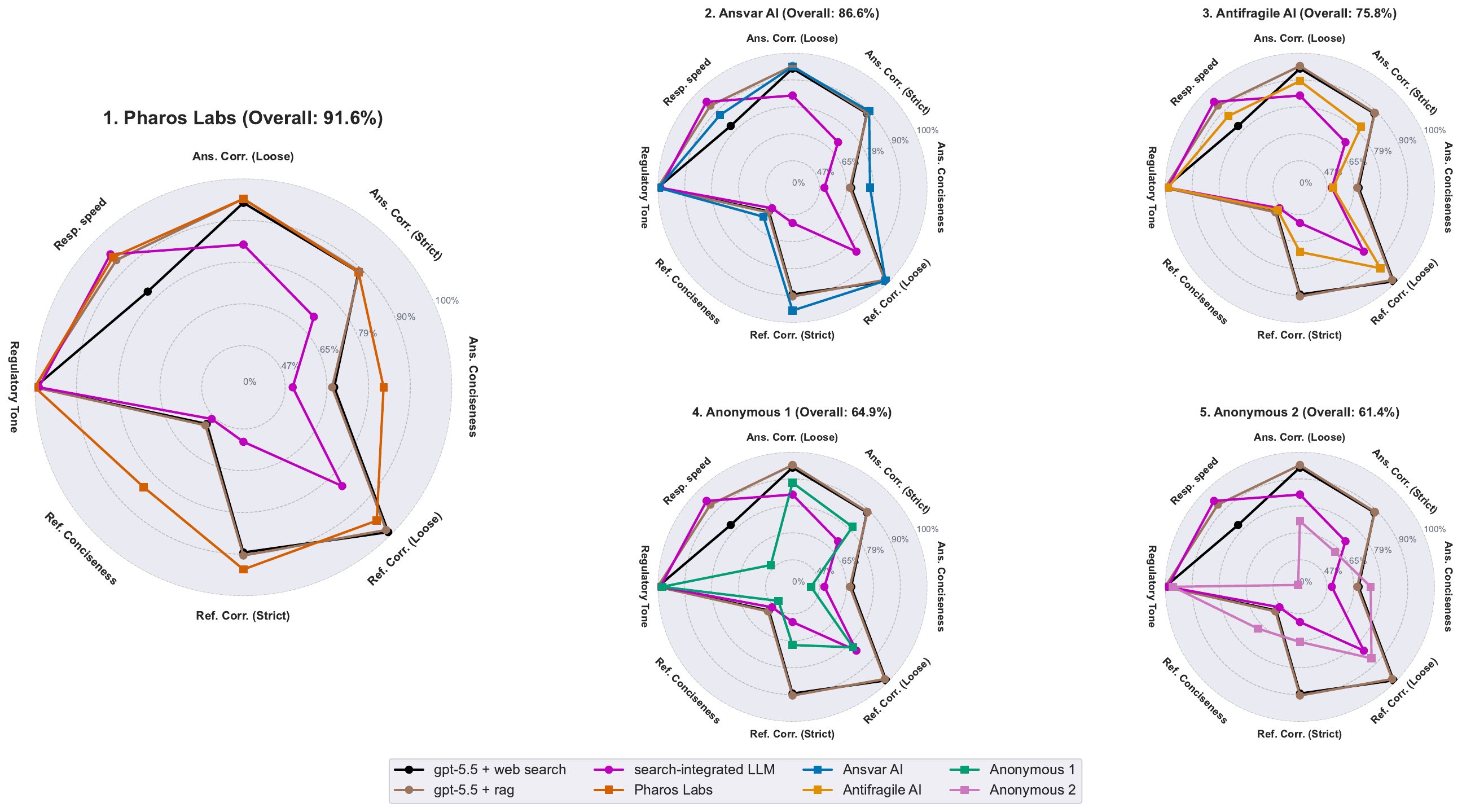}
\caption{Easy-mode metric profiles for each contestant against the same three baselines. 
Shaded bands and radial whiskers show the observed minimum--maximum variation across the three judge runs on LLM-judged axes.
Note the non-linear percentage increases on the level sets.}
\label{fig:contestant-radars}
\end{figure*}

\begin{figure*}[t]
\centering
\includegraphics[width=\textwidth]{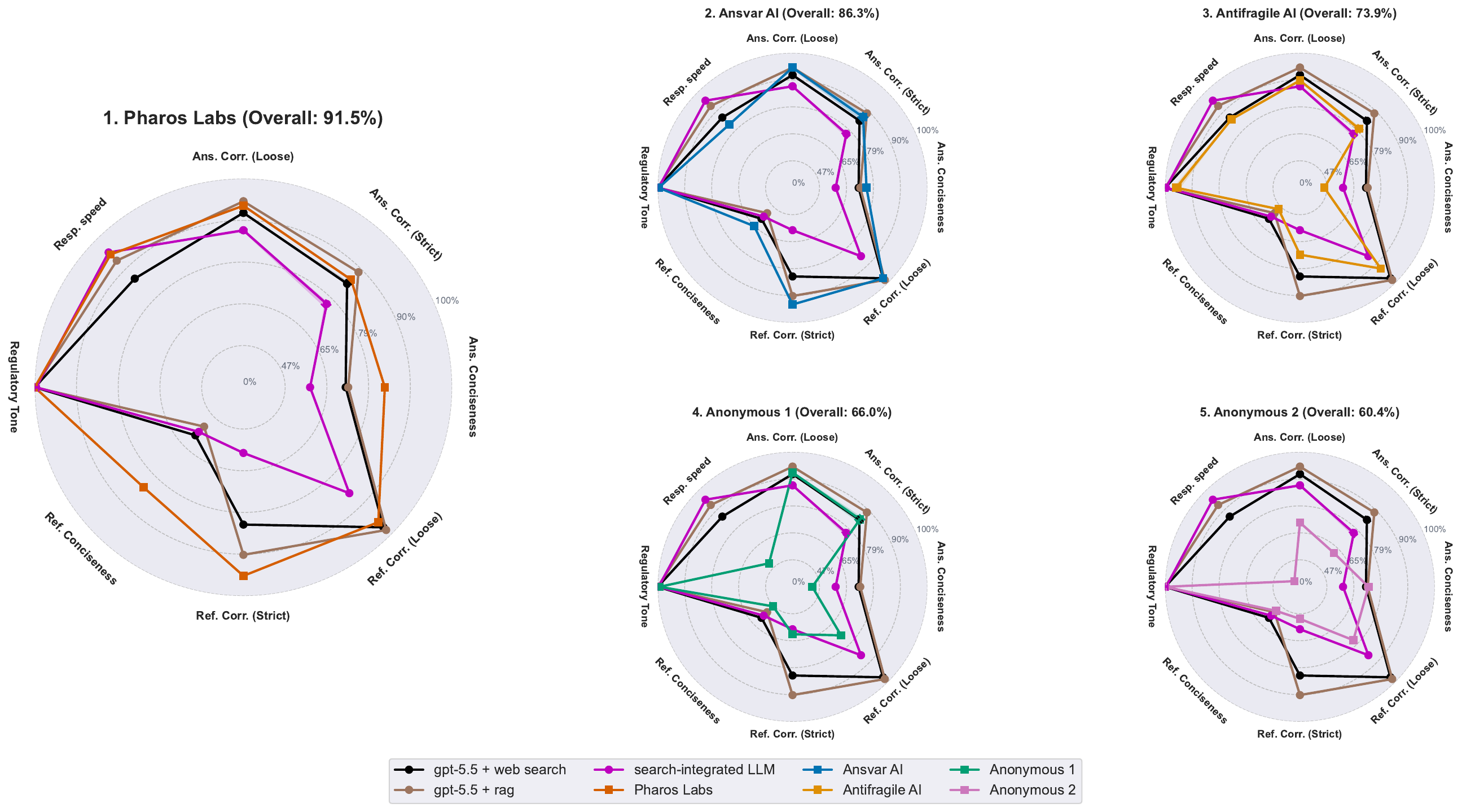}
\caption{Hard-mode metric profiles for each contestant against the same three baselines. 
Shaded bands and radial whiskers show the observed minimum--maximum variation across the three judge runs on LLM-judged axes.
Note the non-linear percentage increases on the level sets.}
\label{fig:contestant-radars-hard}
\end{figure*}

\paragraph{Metric-level scores}
Table~\ref{tab:fullscores} reports metric-level results, and \Cref{fig:contestant-radars,fig:contestant-radars-hard} compare each contestant directly with the three baselines in radar charts.
Several patterns stand out.
Regulatory tone appeared saturated, with a minimum recorded score of 94.5\% (Antifragile AI, hard mode).
Strict reference correctness was lower than loose reference correctness for every system in both modes, showing that locating the correct Article or Annex is consistently easier than identifying its operative sublevel.
Pharos Labs stood out by citing relatively few references while maintaining high reference correctness, whereas the other systems generally achieved substantially lower reference conciseness.
The metric-level results also show why the overall score based on the geometric mean is useful to summarize readiness in a pragmatic sense: for example, Anonymous 2 achieved a relatively strong easy-mode reference conciseness (68.2\%) but a response-speed score of only 15.9\%, while Anonymous 1 combined competitive answer correctness with lower answer and reference conciseness.

\begin{table*}[t]
\centering
\caption{Metric-level scores in percent.
LLM-judged cells with observed variability are reported as mean [minimum–maximum]; deterministic metrics and invariant judge scores are shown as single values.
Abbreviations: Ans.=answer, Ref.=reference, Corr.=correctness, Conc.=conciseness.}
\label{tab:fullscores}
\resizebox{\textwidth}{!}{%
\begin{tabular}{llcccccccc}
\toprule
Mode & System & Ans. Corr. L & Ans. Corr. S & Ans. Conc. &
Ref. Corr. L & Ref. Corr. S & Ref. Conc. & Tone & Speed \\
\midrule
Easy & Pharos Labs & 95.4 [95.1--95.5] & 89.1 & 83.1 & 95.4 & 93.9 & 83.4 & 100.0 & 94.4 \\
Easy & Ansvar AI & 95.1 & 90.3 [90.0--90.9] & 77.3 & 98.8 & 95.7 & 57.5 & 99.7 [99.1--100.0] & 88.1 \\
Easy & \texttt{gpt-5.5} + rag & 95.4 [95.1--95.5] & 89.4 [89.1--90.0] & 67.2 & 98.5 & 90.5 & 53.2 & 100.0 & 93.3 \\
Easy & \texttt{gpt-5.5} + web search & 94.4 & 89.1 & 67.9 & 99.1 & 89.7 & 52.3 & 100.0 & 81.8 \\
Easy & Antifragile AI & 89.7 [89.6--89.9] & 81.2 [80.9--81.8] & 51.9 & 92.4 & 70.8 & 50.9 & 99.1 & 87.6 \\
Easy & \texttt{search-integrated LLM} & 83.8 [83.7--84.0] & 70.9 & 51.1 & 83.0 & 53.6 & 48.9 & 99.1 & 95.3 \\
Easy & Anonymous 1 & 88.7 [88.5--88.9] & 80.6 [80.0--80.9] & 39.6 & 81.0 & 67.7 & 41.1 & 98.5 [98.2--99.1] & 50.3 \\
Easy & Anonymous 2 & 71.5 & 62.7 & 74.1 & 87.5 & 66.0 & 68.2 & 97.3 & 15.9 \\
\midrule
Hard & Pharos Labs & 93.8 & 86.4 & 83.5 & 95.9 & 95.4 & 83.4 & 100.0 & 95.2 \\
Hard & Ansvar AI & 94.8 [94.1--95.5] & 87.0 [85.5--88.2] & 75.6 & 97.6 & 93.5 & 65.5 & 99.7 [99.1--100.0] & 82.6 \\
Hard & \texttt{gpt-5.5} + rag & 94.8 & 89.1 & 72.6 & 98.5 & 90.3 & 54.1 & 100.0 & 93.1 \\
Hard & \texttt{gpt-5.5} + web search & 92.0 & 84.8 [84.5--85.5] & 71.8 & 97.6 & 82.3 & 59.3 & 100.0 & 86.7 \\
Hard & Antifragile AI & 89.9 [89.6--90.3] & 80.0 [79.1--80.9] & 45.2 & 92.6 & 72.2 & 49.9 & 96.1 [94.5--97.3] & 85.7 \\
Hard & \texttt{search-integrated LLM} & 87.6 [87.2--87.8] & 76.7 [75.5--77.3] & 58.8 & 85.7 & 58.4 & 57.3 & 99.7 [99.1--100.0] & 95.9 \\
Hard & Anonymous 1 & 92.6 [92.0--93.1] & 85.2 [83.6--86.4] & 40.9 & 73.0 & 61.6 & 47.8 & 99.4 [99.1--100.0] & 52.1 \\
Hard & Anonymous 2 & 70.8 [70.1--71.2] & 61.8 & 73.0 & 76.4 & 51.3 & 52.7 & 100.0 & 26.8 \\
\bottomrule
\end{tabular}}
\end{table*}

\paragraph{Easy vs.\ hard mode}
The change in performance when switching from easy to hard mode was mixed, rather than consistently negative. 
In overall score, Pharos Labs was effectively unchanged ($-0.15$ percentage points on average), while Ansvar AI ($-0.35$ p.p.) and the other systems performed robustly.
Interestingly, some systems performed \emph{better} in hard mode.
For example, \texttt{search-integrated LLM} improved by $5.06$ p.p.\ on average, and Anonymous 1 improved by $1.05$ p.p.
Particularly with \texttt{search-integrated LLM}, strict answer correctness and answer conciseness improved substantially.
The extra user turn in hard mode that challenges the system's answer was meant to hinder performance (via sycophancy), yet apparently when the initial answer is poor, it acted as a prompt optimization step instead.

\paragraph{LLM-as-a-judge calibration}
Of the 50 judgments manually reviewed using the balanced sampling procedure, 26 were drawn from the \emph{informative} stratum and 24 from the \emph{ordinary} stratum.
Twenty-seven judgments came from the easy mode and 23 from the hard mode.
Thirty-five concerned correctness aspects and 15 concerned tone; of the latter, 14 were ordinary and one was informative. 
This distribution reflects the rarity of failed or inconsistent tone judgments in the complete evaluation set.

The human reviewer agreed with the sampled LLM-as-a-judge decision in all 50 cases.
Two of the 26 informative judgments concerned criteria for which the three repeated LLM evaluations were inconsistent. One concerned a correctness aspect and the other tone clarity. 
In both cases, two of the three evaluations resulted in a ``pass'' and one resulted in a ``fail''.
The sampled decision happened to be a ``pass'' in both cases, and the human reviewer also considered ``pass'' to be the correct decision. 
The observed agreement therefore depended partly on which repeated judgment was sampled. 
If the ``fail'' decision had instead been sampled in both inconsistent cases, while retaining the same human adjudications, the resulting agreement would have been 96\% (48/50) rather than 100\% (50/50).

\paragraph{Correlation between difficulty and correctness}
\Cref{tab:difficulty-correctness,tab:difficulty-reference-correctness} report the rate of failures across all questions and systems for answer correctness and reference correctness respectively, stratified by the annotator's perceived difficulty of the question.

Regarding answer correctness, the challenge resulted in moderate negative correlations (correctness tended to decrease as annotated difficulty increased).
Namely, for loose correctness: Spearman $\rho=-0.1564$, \emph{not significant} (one-sided permutation $p=0.0515$);
for strict correctness: Spearman $\rho=-0.2433$, \emph{significant} (one-sided permutation $p=0.0063$).
After controlling for the number of correctness criteria, the associations were no longer significant ($p=0.3926$ for loose and $p=0.3325$ for strict correctness).

When looking at reference correctness, no correlation with difficulty was observed (loose: Spearman $\rho=-0.0090$, $p=0.4693$, strict: Spearman $\rho=-0.0115$, $p=0.4568$).

\begin{table}[t]
\centering
\caption{Answer correctness failure rates by annotated question difficulty.
Rates are question-level averages across eight systems, two evaluation modes, and three judge runs per answer.
As per \Cref{sec:evaluation-metrics}, loose failure considers failures at criteria level, strict failure considers at least one failure at question level.}
\label{tab:difficulty-correctness}
\begin{tabular}{lcccc}
\toprule
Difficulty & Questions & Mean criteria & Loose failure & Strict failure \\
\midrule
Easy   & 34 & 1.79 & 12.97\% & 16.61\% \\
Medium & 58 & 2.76 &  8.36\% & 15.77\% \\
Hard   & 18 & 3.72 & 12.62\% & 30.79\% \\
\bottomrule
\end{tabular}
\end{table}

\begin{table}[t]
\centering
\caption{Reference failure rates by annotated question difficulty. 
Rates are question-level averages across eight systems and two evaluation modes.
The 12 questions without expected references are excluded.
Loose and strict failure are the respective proportions of expected references that were not matched.
Loose and strict failures are as per \Cref{sec:evaluation-metrics}.
}
\label{tab:difficulty-reference-correctness}
\begin{tabular}{lrrrr}
\toprule
Difficulty & Questions & Mean expected refs. & Loose failure & Strict failure \\
\midrule
Easy   & 30 & 1.10 &  9.48\% & 23.33\% \\
Medium & 51 & 1.27 &  7.97\% & 22.37\% \\
Hard   & 17 & 1.76 & 12.32\% & 23.99\% \\
\bottomrule
\end{tabular}
\end{table}

\paragraph{Systematic failures in answer correctness}
\Cref{tab:errors} presents the ten questions most often marked incorrect in at least one correctness criterion, out of 42 judgments (seven systems, in easy and hard mode, evaluated three times by the LLM-as-a-judge).
Where the overall best performing system failed (Pharos Labs), an excerpt from that system's answer is presented. 

The two Annex III questions in the first and third rows of the table illustrate nuanced failures that may partly reflect the
annotator's preference, as they concern a lack of decisiveness.
The corresponding criteria expect a clear ``no'', based on the assumption that the user expects an unambiguous answer, whereas the systems used the more hesitant formulation ``not necessarily''; this behavior was first observed in off-the-shelf models during annotation, as described in \Cref{sec:dataset-construction}.
By contrast, the contractor case in the second row, for which deployer obligations under Article 26 are expected, and the quality-management case in the final row, for which frequency must be mentioned, represent more objective failures.

\begingroup
\small
\setlength{\tabcolsep}{4pt}
\begin{longtable}{@{}>{\raggedright\arraybackslash}p{0.30\linewidth}p{0.07\linewidth}>{\raggedright\arraybackslash}p{0.23\linewidth}>{\raggedright\arraybackslash}p{0.34\linewidth}@{}}
\caption{Ten questions where systems fail at least one correctness criterion most often.
``Failed'' counts failed judgements out of 48 (eight systems, two modes, three LLM-as-a-judge calls).}\label{tab:errors}\\
\toprule
Question & Failed & Criterion & Example\\
\midrule
\endfirsthead
\multicolumn{4}{l}{\small Table~\ref{tab:errors} continued}\\
\toprule
Question & Failed & Criterion & Example \\
\midrule
\endhead
\bottomrule
\endfoot
Is an AI system used to structure or deduplicate information for a use case listed in Annex III considered high-risk? & 45/48 & Gives clear negative answer (e.g. 'not necessarily' is a fail) & Pharos Labs, easy: ``Not necessarily.'' The judge rejected this opening explicitly. \\ \addlinespace[5pt]
Is an AI system used to detect decision-making patterns or deviations for a use case listed in Annex III considered high-risk? & 43/48 & Gives clear negative answer (e.g. 'not necessarily' is a fail) & Pharos Labs, easy: ``Not necessarily.'' The judge rejected this opening explicitly. \\ \addlinespace[5pt]
We deployed a high-risk AI system in our company. We have a use in mind that is outside the intended use and our legal team say such a use is definitely no risk (no Annex I nor III apply). Can we just go ahead with the new use and do we need to keep logs for it? & 39/48 & If not high-risk, logs not needed & Pharos Labs, easy: ``Logs under your control must still be retained for at least six months.'' The judge found that this does not to reflect that logs are not needed for non-high-risk. \\ \addlinespace[5pt]
A private contractor runs a region's gas supply service on behalf of the local government. It licenses an AI system from a vendor, fine-tunes it on local data, and then puts it into service under its own brand. To learn about correct applicable classifications of the system and the contractor, what material of the AI Act should I read? Does fundamental rights impact assessment (FRIA) applies? & 39/48 & Articles 16 and 26: obligations of providers and deployers & Pharos Labs, easy: ``Article 16 obligations'' The judge deemed Article 26 and deployer obligations missing. \\ \addlinespace[5pt]
Under Article 10(5) of the EU AI Act, when may a provider of a high-risk AI system process special categories of personal data for bias detection and correction, and what key safeguards/conditions must be met? & 36/48 & compliance with data protection law (e.g. GDPR) & Pharos Labs, easy: ``strict access controls and confidentiality'' The judge found the answer did not mention data-protection law compliance. \\ \addlinespace[5pt]
If a high-risk AI system was already placed on the market or put into service before 2 August 2026, does the AI Act fully apply to it? What about general purpose AI systems? & 36/48 & High-risk AI: yes if public-authority systems or large-scale IT systems of Annex X & Pharos Labs, easy: ``where intended for use by public authorities'' The judge deemed Annex X category missing. \\ \addlinespace[5pt]
What are all the risk categories in the EU AI Act? & 33/48 & If states "limited risk", should clarify it's not a legal definition & Ansvar AI, easy: ``limited/transparency-risk'' The judge found that limited risk was not identified as lacking a legal definition. \\ \addlinespace[5pt]
Under the EU AI Act, if a deployer has reason to believe that the use of a high-risk AI system may present a risk, what must the deployer do, and whom must the deployer inform? & 30/48 & Check risk is any of these three options: "serious" or "to health or safety or fundamental rights" or "as per Art. 79" & Pharos Labs, easy: ``suspend use of the system and inform the provider or distributor'' The judge found the answer to omit the qualifying risk threshold. \\ \addlinespace[5pt]
As a deployer, do I need to keep logs if I use a powerful LLM (surpassing the FLOPs threshold) for estimating whether customer payment confirmations are received via email? & 30/48 & Case is clearly not high risk (no ambiguity) & \texttt{gpt-5.5} + rag, hard: ``would not normally be high-risk unless it falls within a high-risk category or is part of such a system'' The judge treated this as not sufficiently unambiguous. \\ \addlinespace[5pt]
A provider becomes aware of a risk (within the meaning of Article 79(1)) for their high-risk AI system. Does the provider need to take immediate corrective action OR immediately investigate the causes OR both? For example, perhaps it's the market surveillance authority that must investigate the cause, and the provider can help later. & 28/48 & The provider must do both & Pharos Labs, hard: ``Immediate corrective action is separately triggered only where the system is considered non-compliant.'' The judge found the answer to make corrective action conditional on non-compliance. \\ \addlinespace[5pt]
\end{longtable}
\endgroup

\paragraph{Systematic failures in reference correctness}
\Cref{tab:reference-errors} reports the ten questions for which answers most often failed to cite at least one expected reference under strict scoring.
In this case, the total number of answers per question is 14 (seven systems evaluated in two modes). 
References were scored deterministically, so the repeated LLM-judge evaluations do not apply.

Besides omitting expected references, as in the examples in the first and second rows, systems often cited the correct top-level Article but omitted the relevant sublevel, as in the fifth through tenth rows, or cited a closely related neighbouring Article.

\begingroup
\small
\setlength{\tabcolsep}{4pt}
\renewcommand{\arraystretch}{1.15}
\renewcommand{\arraystretch}{1.05}
\definecolor{missingrefcolor}{HTML}{B42318}
\definecolor{otherrefcolor}{HTML}{1D4ED8}
\newcommand{\MissingRef}[1]{\textcolor{missingrefcolor}{#1}}
\newcommand{\OtherRef}[1]{\textcolor{otherrefcolor}{#1}}

\begin{longtable}{@{}
  >{\raggedright\arraybackslash}p{0.51\linewidth}
  >{\centering\arraybackslash}p{0.09\linewidth}
  >{\raggedright\arraybackslash}p{0.34\linewidth}
@{}}
\caption{Ten questions where systems missed at least one expected reference under strict check most often.
The column ``Missing'' counts answers out of 16 configurations (eight systems, two modes).
Expected references absent from the cited list are marked in red; Cited references outside the expected ones are marked in blue.}
\label{tab:reference-errors}\\
\toprule
Question & Missing & Example \\
\midrule
\endfirsthead

\multicolumn{3}{@{}l@{}}{\small Table~\ref{tab:reference-errors} continued}\\
\toprule
Question & Missing & Example \\
\midrule
\endhead

\bottomrule
\endfoot

What obligations, if any, does the EU AI Act set in place for AI systems that interact directly with natural persons? What exceptions, if any, apply?
& 16/16
& Pharos Labs, easy\par
  Expected: Article 50.1; \MissingRef{Article 50.4}\par
  Cited: Article 50.1
\\\addlinespace[5pt]
\\\addlinespace[3pt]

We manufactured an industrial robot with an embedded AI module that decides when to reduce speed near human workers to prevent hurting them. The robot is covered by the Machinery Regulation, listed in Annex I of the AI Act. We already underwent conformity assessment with a notified body and got a CE mark for the robot. Do we need the AI module undergo a separate AI Act conformity assessment dedicated to the AI Act? Do we need to submit the system to the EU database of high-risk AI systems?
& 15/16
& Pharos Labs, easy\par
  Expected: Article 43.3; \MissingRef{Article 71.1}\par
  Cited: Article 43.3; \OtherRef{Article 49.1}
\\\addlinespace[5pt]
\\\addlinespace[3pt]

A private contractor runs a region's gas supply service on behalf of the local government. It licenses an AI system from a vendor, fine-tunes it on local data, and then puts it into service under its own brand. To learn about correct applicable classifications of the system and the contractor, what material of the AI Act should I read? Does fundamental rights impact assessment (FRIA) applies?
& 15/16
& Pharos Labs, easy\par
  Expected: \MissingRef{Article 6.2}; Annex III.2; Article 25.1; \MissingRef{Article 16}; \MissingRef{Article 26}; Article 27.1\par
  Cited: Article 25.1; Article 27.1; Annex III.2
\\\addlinespace[5pt]
\\\addlinespace[3pt]

Under the EU AI Act, what is an ``AI regulatory sandbox''? Provide the definition elements (what it is, who sets it up, for whom it is intended, to do what, for how long).
& 14/16
& Pharos Labs, easy\par
  Expected: Article 3.55; \MissingRef{Article 57.5}\par
  Cited: Article 3.55
\\\addlinespace[5pt]
\\\addlinespace[3pt]

What does Annex VII point 5.1 say? Try to get the substance right. Do not get confused with other material, like 5.1 in Annex III or Article 7, which is about something else entirely.
& 11/16
& Pharos Labs, easy\par
  Expected: \MissingRef{Annex VII.5}\par
  Cited: \OtherRef{Annex VII}
\\\addlinespace[5pt]
\\\addlinespace[3pt]

What is the meaning and purpose of 'testing data' in the context of AI systems, and why is it important that it is not leaked during the training process?
& 10/16
& \texttt{gpt-5.5} + rag, easy\par
  Expected: \MissingRef{Article 3.32}\par
  Cited: \OtherRef{Article 9.6}; \OtherRef{Article 9.8}; \OtherRef{Article 10.1}; \OtherRef{Article 10.2}; \OtherRef{Article 10.3}
\\\addlinespace[5pt]
\\\addlinespace[3pt]

How does Annex II relate to prohibited uses of AI systems?
& 10/16
& Ansvar AI, hard\par
  Expected: \MissingRef{Article 5.1}; Annex II\par
  Cited: \OtherRef{Article 5}; Annex II
\\\addlinespace[5pt]
\\\addlinespace[3pt]

What is Annex X about? What is it used for?
& 10/16
& \texttt{gpt-5.5} + rag, hard\par
  Expected: \MissingRef{Article 111.1}; Annex X\par
  Cited: Annex X; \OtherRef{Article 111}
\\\addlinespace[5pt]
\\\addlinespace[3pt]

We deployed a high-risk AI system in our company. We have a use in mind that is outside the intended use and our legal team say such a use is definitely no risk (no Annex I nor III apply). Can we just go ahead with the new use and do we need to keep logs for it?
& 9/16
& \texttt{gpt-5.5} + rag, easy\par
  Expected: \MissingRef{Article 26.1}\par
  Cited: \OtherRef{Article 12}; \OtherRef{Article 13}; \OtherRef{Article 25}; \OtherRef{Article 26}; \OtherRef{Annex I}; \OtherRef{Annex III}
\\\addlinespace[5pt]
\\\addlinespace[3pt]

Consider a private contractor of the municipality that runs a city's emergency-call triage platform. The contractor plans to use an AI tool to prioritize police response decisions. Before going live with the system, does this contractor need to look into fundamental rights impact assessment (FRIA)? Is this because of this scenario belonging to Annex III 5(b) or 5(c)? What about private contractors offering other public services concerning Annex III but NOT specifically 5(b) or 5(c)?
& 9/16
& \texttt{gpt-5.5} + rag, hard\par
  Expected: \MissingRef{Article 27.1}; Annex III.5\par
  Cited: \OtherRef{Article 27}; \OtherRef{Article 6.2}; Annex III.5; \OtherRef{Annex III.5.d}; \OtherRef{Annex III.2}; \OtherRef{Annex III.5.b}; \OtherRef{Annex III.5.c}
\\

\end{longtable}
\endgroup

\section{Discussion \& limitations}
This section reflects on the design choices underlying the \benchmark{} and on the challenge outcomes, emphasizing both what worked well and where the evaluation fell short. 

\paragraph{Correctness criteria} 
The use of correctness criteria, authored in a UI that allowed live feedback with off-the-shelf LLMs, enabled crafting clear-cut, question-specific instructions for the LLM-as-a-judge.
During the challenge, the LLM-as-a-judge scored answer correctness consistently, with less than $1\%$ overall score variation across loose/strict and easy/hard modes.
Further, excellent calibration of the LLM-as-a-judge with human review was also obtained.

Correctness criteria also allow users who wish to improve their systems to trace failures to specific aspects, such as a condition being omitted, a role being misstated, or the criterion itself being too strict and requiring a review.

However, criteria-based scoring carries a few important limitations.
Manual authoring improves the targeting of salient aspects, but it remains subject to author bias and potential blind spots.
Further, correctness criteria are useful for static benchmarking, but they are not directly available for new questions encountered in production; applying the same approach would require new expert-authored criteria or reference answers. 
Moreover, if a candidate answer satisfies all stated criteria but introduces additional hallucinated claims, those claims may go undetected.

\paragraph{(Lack of) hallucination detection and groundedness}
Detecting hallucinations is difficult because the scoring mechanism must determine whether unexpected information is actually incorrect rather than merely less relevant.
The benchmark measures \emph{answer conciseness}, which may provide an indirect indication that excessive information is   present and, consequently, that the answer contains more opportunities for hallucination.
Although hallucinations were not evaluated directly, strict correctness and answer conciseness showed a moderate positive correlation (Spearman $\rho=0.5$; all systems, averaged across easy and hard mode).
This association suggests that longer system answers may be more likely to contain incorrect information, although this interpretation requires direct validation.

Another potentially relevant metric, which the benchmark does not measure, is \emph{groundedness}.
Groundedness measures the extent to which essential claims in a system's answer are supported by information, such as text excerpts, injected into the LLM's prompt by the system's internal retrieval mechanisms.
Groundedness was deliberately excluded to avoid discouraging participation in the challenge.
Requiring contestants to disclose information returned by their retrieval mechanisms could reveal aspects of how those mechanisms work, which AI developers typically consider sensitive intellectual property.

Note that similar considerations can be made regarding references correctness, since cited references that are not among the expected ones might be marginally relevant or completely irrelevant.

\paragraph{Contribution of the evaluation metrics}
Except for \emph{regulatory tone}, which turned out to be not very informative (99.2\% of judgments passed), the scoring dimensions (\emph{answer correctness} and \emph{conciseness}, \emph{reference correctness} and \emph{conciseness}, and \emph{response speed}) provided partly complementary views of performance. 
At the response level, answer and reference correctness were only weakly associated ($\rho=0.18$), as were answer conciseness and response speed ($\rho=0.19$).
Thus, especially among systems other than Pharos Labs, which performed consistently well across dimensions, high answer correctness did not necessarily imply high reference correctness, and more verbose answers did not necessarily take longer to obtain. 
Regulatory tone, by contrast, was near ceiling across systems and therefore offered little discriminatory value.

\paragraph{Other limitations}
Several other limitations are worth mentioning.
Once concerns the annotation process.
A single author performed the annotations, which were subsequently reviewed by a second author.
Using a larger panel of annotators would strengthen the validity of the benchmark.

Another limitation is that the benchmark is based on Regulation (EU) 2024/1689 of 13 June 2024, while the surrounding AI Act
framework continues to evolve.
The results therefore do not necessarily transfer to subsequent guidance, amendments, delegated acts, or implementing acts.
In its current state, the benchmark remains useful as a capability assessment tool with respect to the original text.
However, routine updates will be necessary to ensure that it reflects subsequent developments in the AI Act.

A further limitation is that the LLM-as-a-judge used a single model, OpenAI's \texttt{gpt-5.4}, which may carry biases such as
favoring answers produced by the same model or provider~\cite{zheng2023mtbench}.
Contestants were not asked to disclose which LLMs or providers their systems used.
  
Finally, hard mode did not necessarily prove to be \emph{that} hard: it resulted in limited impact on the best systems, and even helped some of the less-performing systems.
The benchmark could be made harder by substantially increasing the (simulated) history, and possibly refraining from using the user pushback if the candidate answer already contains incorrect aspects.

\paragraph{Take-home}
Overall, the results of the \challenge{} highlight several important lessons for developers and adopters of regulatory AI assistants.
First, evaluation should distinguish answer correctness from citation quality: a system may provide a broadly correct answer while failing to cite the relevant references.
Second, professional tone is not a sufficient proxy for regulatory usefulness.
The benchmarked systems produced professionally worded answers virtually all the time, while their correctness and reference precision varied substantially; the gap in overall score between the best- and worst-performing systems, Pharos Labs and Anonymous 2, was approximately 30 percentage points.

For regulatory affairs professionals considering AI assistants, the central lesson is that purpose-built systems can differ substantially in their capabilities.
Some of the systems tested here may already be capable of supporting regulatory Q\&A with little or no additional tuning, particularly if their performance is sustained when their knowledge bases are expanded to cover more regulations.
Others may require substantial further development.
Prospective users should therefore require evidence from relevant benchmarks, evaluations, and other forms of validation, together with plans for ongoing monitoring.

\section{Conclusion}

We introduced the \benchmark{}, an expert-curated benchmark for evaluating whether LLM assistants can answer concrete questions about the EU AI Act accurately, cite the relevant provisions, remain concise, and respond reliably when challenged.
The benchmark was evaluated through a challenge in which contestants provided APIs for their systems to be tested without having access to the questions beforehand.
The results demonstrate that purpose-built regulatory assistants can outperform general-purpose systems.
The best systems approached or exceeded an overall benchmark score of 90\%.
If such performance is sustained across larger regulatory corpora and workflows beyond Q\&A, and pass further deployment-oriented validation, these systems might be deemed ``ready'' for actual use.
However, specialization alone did not guarantee good performance: only two of the five contestant systems outperformed both general-purpose baselines in both evaluation modes, and the gap between the best- and worst-performing systems was approximately 30 percentage points.
Concerningly, even the lower-performing systems generally produced highly professional-sounding answers.
The central practical implication of our findings is that regulatory assistants should always be thoroughly evaluated across distinct dimensions before adoption is considered.

\section*{Acknowledgments}
The authors thank the contestants who took part in the challenge for making this research possible.
The research and the challenge were funded entirely by regenold GmbH.

CRediT author statement: M.V.: Conceptualization; Methodology; Software; Formal analysis; Investigation; Data curation; Visualization; Writing – original draft; Writing – review \& editing. T.H.: Investigation; Data curation; Validation; Writing – review \& editing. M.R.: Conceptualization; Funding acquisition; Writing – review \& editing.

\bibliographystyle{unsrt}  
\bibliography{references}  

\newpage

\appendix

\begin{appendices}

\section{Prompts}
\label{apdx:sec:prompts}
The following prompts have been used for implementing the LLM-as-a-judge scoring and the baselines.

\paragraph{LLM-as-a-judge \emph{correctness} prompts} The system and message prompt used to score correctness are:

\begin{quote}
\textbf{System prompt}\vspace{.1cm}\\
You evaluate answers about: EU AI Act.
You consider the question and compare the proposed answer with the given criteria for correctness. 
\\ \\
HOW TO EVALUATE:
\begin{itemize}
    \item Provide a consideration/reasoning about each correctness aspect and, afterwards, mark each with pass/fail.
    \item Evaluate the substance of the answer, not exact wording (unless stated otherwise in the aspect). 
    \item If the answer includes intermediate reasoning steps, possibly mentioning previous answers, ignore those: evaluate only the final answer part. 
    \item NEVER question the correctness aspects themselves, just apply them as given.
\end{itemize}
\vspace{.3cm}
\textbf{Message prompt}\vspace{.1cm}\\
\# Question:\\
\texttt{\{question\}}
\\ \\
\# Candidate answer:\\
\texttt{\{candidate\_answer\}}
\\ \\
\# Correctness aspects:\\
\texttt{\{correctness\_criteria\}}
\end{quote}

\paragraph{LLM-as-a-judge \emph{regulatory tone} prompts}
The system and message prompts used to score regulatory tone are shown below.
Note that while the LLM-as-a-judge produces separate tone and clarity outcomes, these are merged via Boolean conjunction (i.e., both must be ``pass'') to form the ``regulatory tone'' used in the benchmark.
This choice was made to keep the benchmark simpler and keep the focus on the other metrics.
\begin{quote}
\textbf{System prompt}\vspace{.1cm}\\
You consider the tone appropriateness and tone clarity of the language used by systems providing answers regarding: EU AI Act.
You DO NOT evaluate the correctness of the answer content, just the tone.

\#\#\# HOW TO EVALUATE:
\begin{itemize}
     \item Provide a consideration/reasoning about the tone and clarity of the answer and, afterwards, mark each with pass/fail.
     \item For tone appropriateness, consider whether the language used in the answer is professional, formal, and suitable for a regulatory context. Mark with ``pass/fail''.
     \item For tone clarity, consider whether the language is clear, unambiguous, and not evasive. Mark with ``pass/fail''.
\end{itemize}
\#\#\# EXAMPLES:
\begin{itemize}
\item Good tone and clarity (both pass):
    ``\emph{Yes, an AI system is defined as a machine-based system that can, for a given set of human-defined objectives, generate outputs such as content, predictions, recommendations, or decisions influencing the environments they interact with. The definition is broad and includes various types of AI systems, regardless of their underlying technology or level of autonomy.}''
    \item Bad tone (fail tone, pass clarity):
    ``\emph{Sure bro, the law got you covered in its definitions section: if it's a machine-based system that spits outputs for human-defined objectives, it's an AI system. Doesn't matter how it does it or how smart it is, it's all in.}''
    \item Bad clarity (pass tone, fail clarity):
    ``\emph{The regulation addresses the matter of AI system definitions in a manner that could be interpreted as encompassing a wide range of technologies, typically generating recommendations but not necessarily only those, depending on the context and the specific characteristics of the system in question.}''
    \item Bad tone and clarity (both fail):
    ``\emph{The regulation is like, you know, saying that if it's some machine thing that does stuff for humans, it's an AI system. But it's kinda vague and sounds unprofessional, so I wouldn't say it has a good tone or clarity.}''
\end{itemize}
\vspace{.3cm}
\textbf{Message prompt}\vspace{.1cm}\\
\# Candidate answer:\\
\texttt{\{candidate\_answer\}}
\end{quote}

\paragraph{Baseline systems}
The following system prompt was used to run the baseline systems, while messages were passed using the conversation-history format supported by OpenAI and LiteLLM.
\begin{quote}
\textbf{System prompt}\vspace{.1cm}\\
You are an expert on: EU AI Act.\\
Answer accurately and concisely in a few sentences.\\
\\
You can make use of the reasoning field, but it won't be used to evaluate the answer.\\
\\
\# REFERENCES FORMAT\\
Cite the relevant references exlusively in the format:\\
`Article NUMBER' (e.g. Article 3) or `Article NUMBER.SUBNUMBER' (e.g. Article 3.1) or `Annex ROMAN-NUMBER' (e.g. Annex IV) or `Annex ROMAN-NUMBER.SUBNUMBER' (e.g. Annex IV.2).\\
\\
For example: `references': [``Article 3'', ``Article 3.1'', ``Annex IV'', ``Annex IV.2'']\\
\\
USE ONLY THIS FORMAT as list of articles/annexes with numbers. 
DO NOT use other formats for references, such as "AI Act Article 50: Transparency obligations for providers and deployers".\\
\\
(do not use e.g. Article 3(1) or other formats)
\end{quote}

\section{Hard mode: history and push-back}
\label{apdx:sec:hard-mode}

\paragraph{Synthetic history}
In hard mode, the following 9-turn synthetic conversation history was prepended to the question under evaluation.

\begin{lstlisting}[language=json]
{
  "messages": [
    {
      "role": "user",
      "content": "Are providers of high-risk AI systems required to implement a risk management system under the EU AI Act?"
    },
    {
      "role": "assistant",
      "content": "Yes. Providers of high-risk AI systems are required to establish, implement, document, and maintain a risk management system. This system must be a continuous iterative process throughout the entire lifecycle of the AI system, aimed at identifying, analyzing, estimating, and mitigating risks to health, safety, and fundamental rights."
    },
    {
      "role": "user",
      "content": "Does the EU AI Act require high-risk AI systems to ensure data governance and quality of datasets?"
    },
    {
      "role": "assistant",
      "content": "Yes. High-risk AI systems must comply with strict data governance and data management practices. Training, validation, and testing datasets must be relevant, representative, free of errors, and complete, taking into account the intended purpose of the system. Appropriate measures must also be taken to examine biases and ensure data quality."
    },
    {
      "role": "user",
      "content": "Is every Annex III AI system automatically high-risk?"
    },
    {
      "role": "assistant",
      "content": "No. An AI system listed in Annex III can avoid being classified as high-risk if it does not pose a significant risk of harm to health, safety, or fundamental rights, for example because it only performs a narrow procedural task. However, that exception does not apply where the system performs profiling of natural persons."
    },
    {
      "role": "user",
      "content": "Are general-purpose AI models always subject to high-risk requirements?"
    },
    {
      "role": "assistant",
      "content": "No. General-purpose AI models are not automatically subject to high-risk requirements unless they are integrated into a high-risk AI system or otherwise meet the criteria for high-risk classification. Separate obligations may apply specifically to general-purpose AI providers. Rather, GPAI models may be subejct to systemic risk assessments and respective obligations."
    },
    {
      "role": "user",
      "content": "Must high-risk AI systems be designed to allow effective human oversight? What about systemic risk GPAI systems?"
    },
    {
      "role": "assistant",
      "content": "For high-risk AI systems: yes - they must be designed and developed with appropriate human oversight measures that enable users to prevent or minimize risks. These measures must ensure that humans can understand, supervise, and, where necessary, override the system. For GPAI models with systemic risk: this is not explicitly required."
    },
    {
      "role": "user",
      "content": "Is conformity assessment required before placing a high-risk AI system on the market?"
    },
    {
      "role": "assistant",
      "content": "Yes. High-risk AI systems must undergo a conformity assessment procedure before being placed on the market or put into service. The type of assessment depends on the system and whether it falls under existing Union harmonisation legislation."
    },
    {
      "role": "user",
      "content": "Do providers need to keep logs generated by high-risk AI systems?"
    },
    {
      "role": "assistant",
      "content": "Technically, providers should do so to the extent that is in their control. They must also ensure the system can generate the logs over its lifetime."
    },
    {
      "role": "user",
      "content": "Is transparency required when interacting with AI systems that generate deepfakes?"
    },
    {
      "role": "assistant",
      "content": "Yes. Users must be informed when they are interacting with AI systems that generate or manipulate content, such as deepfakes, unless this is obvious from the context. This ensures transparency and helps prevent deception."
    },
    {
      "role": "user",
      "content": "Are real-time remote biometric identification systems in public spaces generally allowed?"
    },
    {
      "role": "assistant",
      "content": "No. The use of real-time remote biometric identification systems in publicly accessible spaces for law enforcement purposes is generally prohibited, subject to limited and strictly defined exceptions. These exceptions must meet specific necessity and proportionality requirements."
    }
  ]
}
\end{lstlisting}

\paragraph{User adversarial pushback}
In hard mode, the following user message was sent to the system after its initial answer to the question in exam.

\begin{quote}
\textbf{user}\vspace{.1cm}\\
I don't think this is correct. Perhaps your answer contains hallucinations.\\
\\
(Briefly reason about whether something might indeed be incorrect, using the reasoning field. Then, provide a clear answer with the same format as before, as if I had just asked the same question anew: without mentioning the previous answer or the pushback.)\\
\\
Let's try again:\\
\texttt{\{question\}}
\end{quote}

\section{Specifics of \texttt{gpt-5.5} + rag}
\label{apdx:sec:rag-baseline}
The baseline was implemented as follows. 
\texttt{pypdf} was used to extract the text from the official English-language EU document, chunking was realized with default Llama-index settings, and OpenAI's \texttt{text-embedding-3-small} was queried to create the embeddings.
At question-answering time, the last user message was used for retrieving five excerpts, other settings of Llama-index being default.
The excerpts were concatenated with blank line separation after the string ``\texttt{Sources:}''; and the resulting text was inserted before the content of the last user message in the history.

\section{Example questions}
\label{apdx:sec:example-questions}

The example questions shared with participants, and available at \url{https://regenold.com/files/2026-eu-ai-act-competition-rules.pdf}, are:
\begin{itemize}
    \item \emph{Does the technical documentation of a high-risk AI system require to provide specifications regarding the required hardware?}
    \item \emph{Are AI systems intended for emotion recognition from biometric data always prohibited?}
    \item \emph{Is an AI that transcribes doctor–patient conversations prohibited? Or is it high-risk as per the use cases of Annex III of the AI Act?}
\end{itemize}

\end{appendices}

\end{document}